\documentclass[a4paper]{article}

\usepackage{INTERSPEECH2021}
\usepackage{hyperref}
\usepackage{subcaption}

\title{A Comparative Study on Neural Architectures and Training Methods for Japanese Speech Recognition}
\name{Shigeki Karita, Yotaro Kubo, Michiel Adriaan Unico Bacchiani, Llion Jones}
\address{Google}
\email{\{karita,yotaro,michiel,llion\}@google.com}

\begin{document}

\maketitle

\begin{abstract}
End-to-end (E2E) modeling is advantageous for automatic speech recognition (ASR) especially for Japanese since word-based tokenization of Japanese is not trivial, and E2E modeling is able to model character sequences directly. This paper focuses on the latest E2E modeling techniques, and investigates their performances on character-based Japanese ASR by conducting comparative experiments. The results are analyzed and discussed in order to understand the relative advantages of long short-term memory (LSTM), and Conformer models in combination with connectionist temporal classification, transducer, and attention-based loss functions. Furthermore, the paper investigates on effectivity of the recent training techniques such as data augmentation (SpecAugment), variational noise injection, and exponential moving average. The best configuration found in the paper achieved the state-of-the-art character error rates of 4.1\%, 3.2\%, and 3.5\% for Corpus of Spontaneous Japanese (CSJ) eval1, eval2, and eval3 tasks, respectively. The system is also shown to be computationally efficient thanks to the efficiency of Conformer transducers.
\end{abstract}

\section{Introduction}
End-to-end (E2E) models greatly simplify Japanese automatic speech recognition (ASR) systems. 
Japanese texts have no word boundaries (e.g., white space in English), and therefore it requires a word segmenter for being represented in hidden Markov models (HMMs) and pronunciation dictionaries.
On the other hand, E2E models can directly learn and transcribe raw Japanese texts as character sequences from speech features.

The first Japanese E2E model~\cite{Hori2017} that reported a lower character error rate (CER) than the HMM based system employs a encoder network with convolutional and bidirectional long short-term memory~\cite{lstm} (BLSTM) layers.
Some previous works substitute the encoder network to improve the ASR performance.
Transformer~\cite{transformer} successfully reduces CERs by replacing the BLSTM on Japanese ASR tasks~\cite{karita2019}.
Its successor with several modifications for ASR, Conformer~\cite{Gulati2020}, further decreases CERs on the Japanese tasks as well as other languages~\cite{guo2020recent}.

Most E2E Japanese ASR systems~\cite{Hori2017,watanabe2018espnet,Moriya2020,ueno8462576} employ connectionist temporal classification (CTC)~\cite{ctc} and attention~\cite{Chorowski2014} decoders. For example, \cite{Hori2017}  proposes training and decoding methods using hybrid CTC-attention algorithms. In contrast, transducer decoder~\cite{rnnt}
is not widely used for Japanese ASR. However, the transducer decoder is important for practical applications~\cite{Li2020}. This is because transducer decoders are suitable for streaming applications like CTC decoders, while it can learn dependencies between output tokens as in attention decoders.

In addition to the choice of network architectures, training methods are critical to achieve good ASR performance. For example, SpecAugment~\cite{SpecAugment} significantly improves ASR accuracy in many tasks~\cite{karita2019}. Furthermore, variational noise injection~\cite{vn}, and exponential moving averaging of parameters~\cite{zhang2020pushing} are shown to be effective for training ASR models. Although those methods have been widely used, detailed comparative experiments have not been conducted yet.

Our main contribution in this paper is a detailed comparison of the various encoder-decoder architectures and training methods in Japanese ASR. Combinations of the latest methods as described above are compared over the same task for better understanding their characteristics.
In our experiments, we evaluate their performance in multiple aspects. For example, we measure character error rate, training throughput, convergence, and inference real-time factor on Corpus of Spontaneous Japanese (CSJ).
 
\section{Neural network architectures}

Our ASR models consist of two subnetworks named encoder and decoder. The encoder $\textrm{Enc}(\cdot)$ transforms a sequence of log-mel filterbank speech features $X_\textrm{fbank}$ into a sequence of encoded vectors $X_\textrm{enc} = \textrm{Enc}(X_\textrm{fbank})$.
The decoder receives $X_\textrm{enc}$ and a sequence of token ids $Y$ to predict a posterior distribution $p(Y|X_\textrm{enc}) = \textrm{Dec}(X_\textrm{enc}, Y)$.
During training, the encoder and decoder jointly learn to maximize posterior probability of the correct target sequence.

\subsection{BLSTM encoder}

The most widely used architecture for the encoders  $\textrm{Enc}(\cdot)$ are based on stacked convolution and BLSTM layers~\cite{watanabe2018espnet,Hori2017,Chorowski2014,SpecAugment}.
For faster processing in the following modules, this architecture first subsamples the input features $X_\textrm{fbank}$ by convolution layers with non-linear activation.
Following BLSTM layers recursively transform the subsampled features frame-by-frame.
We refer to this encoder as \textit{BLSTM encoder}.

Since computation of BLSTMs are known to be difficult to parallelize~\cite{DBLP:journals/corr/AppleyardKB16}, it is also difficult to fully utilize accelerators, such as GPUs and TPUs.

\subsection{Conformer encoder}

Conformer is a recently proposed architecture for encoders~\cite{Gulati2020}. To model global and local dependencies in speech features, \textit{Conformer encoder} employs self-attention~\cite{transformer} and convolution modules instead of the BLSTM layers.

\begin{figure}[tb]
    \centering
    \includegraphics[width=0.95\linewidth]{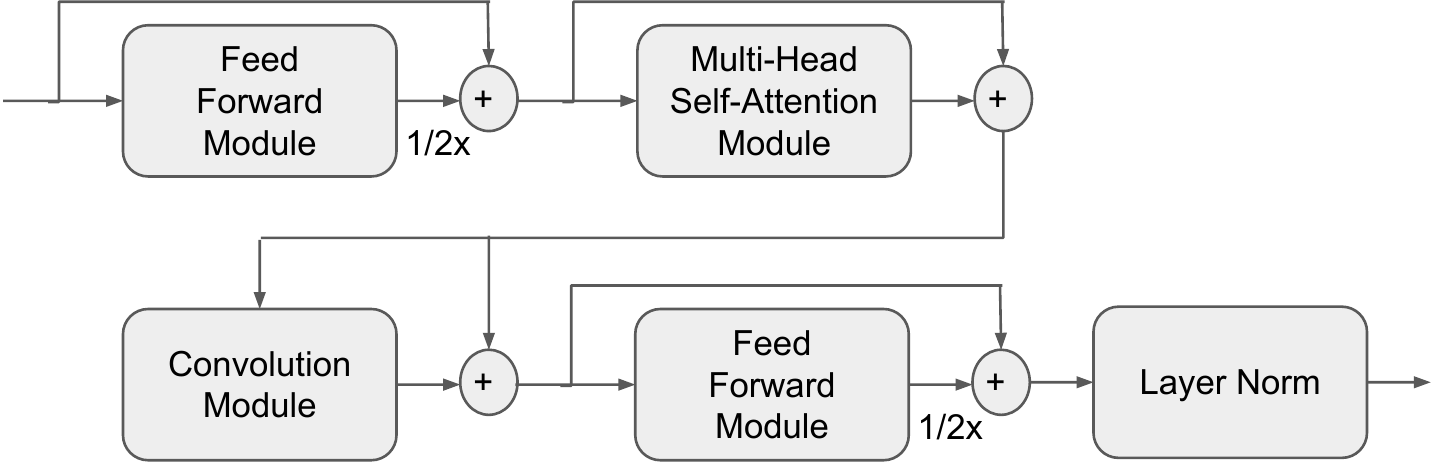}
     \vspace{-0.2cm}
\caption{Conformer encoder block.}
     \label{fig:conformer-block}
     \vspace{-0.2cm}
\end{figure}

Figure~\ref{fig:conformer-block} shows a module diagram of a Conformer block in our system.
The Conformer encodes the input signal by repeating this block several times.
Multi-head self-attention in the block typically requires $O(T^2)$ computation and memory where $T$ is the length of the input sequence.

\begin{figure}[tb]
     \centering
    \includegraphics[width=0.9\linewidth]{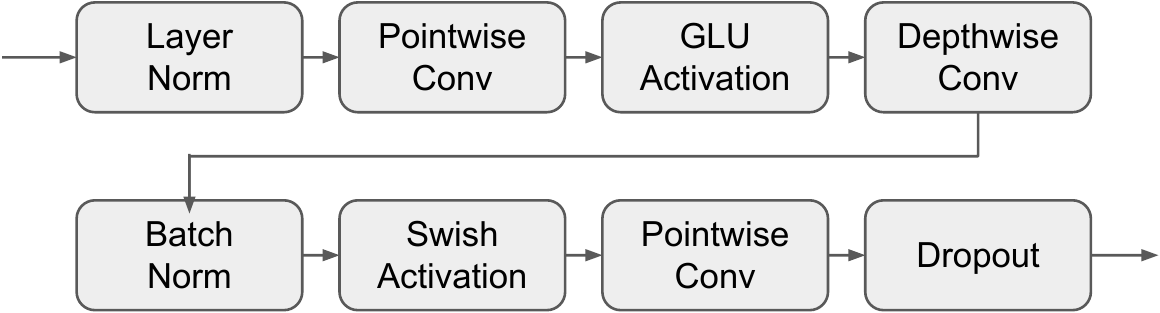}
     \vspace{-0.2cm}
    \caption{Conformer convolution module.}
    \label{fig:conformer-conv}
    \vspace{-0.5cm}
\end{figure}

Figure~\ref{fig:conformer-conv} illustrates the convolution module in a Conformer block.
The convolution module models local features, and consists of layer normalization, first pointwise convolution, gated linear unit (GLU) activation~\cite{DBLP:journals/corr/DauphinFAG16}, 1D depthwise convolution, batch normalization, Swish activation~\cite{swish}, second pointwise convolution, and dropout with a residual connection. In addition, Conformers adopt relative positional encoding~\cite{dai-etal-2019-transformer} in multi-head self attention for modeling global features.

Unlike BLSTM encoders, Conformer encoder does not include recurrent connection, which would disable efficient parallel computation.
Therefore, training with Conformer encoders are expected to be faster than that of BLSTM encoders.
However, as we mentioned above, self-attention block in the Conformer encoders might be a performance bottleneck when the input sequence is long.
In the experimental section, we will compare them in terms of training throughput.


\subsection{Connectionist-temporal-classification decoder}

Our CTC~\cite{ctc} decoder consists of one learnable affine transformation followed by softmax activation.
It predicts the distribution of target sequences $p_\mathrm{ctc}(Y|X_\mathrm{enc})$ with introducing an alignment variable $Z$ that shows correspondence between elements in the encoder output $X_\mathrm{enc}$ and the target sequence $Y$. 

The relation between the alignment variable $Z$ and the target sequence $Y$ is given by a many-to-one mapping denoted as $Y = \mathcal{B}(Z)$.
In CTC, $\mathcal{B}(Z)$ is defined as removal of blank and repeated symbols, for example,  $\mathcal{B}(\text{aa$\phi$b}) = \text{ab}$, where $\phi$ denotes a blank symbol.

The softmax activation in a CTC decoder defines a probability distribution over $Z$, as follows:
\begin{equation}
\begin{aligned}
    C &= \mathrm{softmax}(X_\mathrm{enc} W_\mathrm{ctc} + b_\mathrm{ctc}),
\end{aligned}
\end{equation}
where $C \in \mathbb{R}^{T \times d_\mathrm{vocab}}$, and an element in $C[t, Z[t]]$ is the probability of the token $Z[t]$ aligned at the $t$-th frame in $X_\mathrm{enc}$. 

The posterior probability and corresponding loss function are defined using the inverse of $\mathcal{B}$.
One-to-many mapping $\mathcal{B}^{-1}$ maps a string into a set of strings with added blank or repeated symbols: $\mathcal{B}^{-1}(Y) = \{ Z | Y = \mathcal{B}(Z) \}$. 
Using $\mathcal{B}^{-1}$, the training loss function for CTC $L_\mathrm{ctc}$ can be expressed as follows:
\begin{equation}
\begin{aligned}
p_\mathrm{ctc}(\mathcal{B}(Z) &= Y | X_\mathrm{enc}) = \prod_{t=1}^{T} C[t, Z[t]], \\
    L_\mathrm{ctc} &= -\log \sum_{Z \in \mathcal{B}^{-1}(Y)} p_\mathrm{ctc}(\mathcal{B}(Z) = Y | X_\mathrm{enc}),
\end{aligned}
\label{eq:ctc}
\end{equation}
where $W_\mathrm{ctc} \in \mathbb{R}^{d_\mathrm{enc} \times d_\mathrm{vocab}}$,  $b_\mathrm{ctc} \in \mathbb{R}^{d_\mathrm{vocab}}$ are decoder parameters.
Even though there is a combinatorial explosion in the number of element in $\mathcal{B}^{-1}(Y)$,  forward-backward algorithm can efficiently compute the summation in the Eq. \eqref{eq:ctc}.

\subsection{Transducer decoder}

Similar to the CTC, transducer~\cite{rnnt} decoder also learns to maximize all the possible alignment probability during training. However, unlike CTC, it does not assume the conditional independency on elements in $Y$.
The transducer decoder typically incorporates recurrent layers to recursively encode a prefix token sequence, and predicts the distributions of alignments.

The transducer decoder models conditional distribution $p_\mathrm{trans}(Z[t] | X_\mathrm{enc}, \mathcal{B}(Z[1:t-1]))$ as similar to Eq. \eqref{eq:ctc} with augmenting the network input with the prefix tokens $Z[1:t-1]$.
Using this probability distribution, the training loss function $L_\mathrm{trans}$ to be maximized is defined as follows:
\begin{equation}
\begin{aligned}
    & L_\mathrm{trans} = \\
    & -\log \sum_{Z \in \mathcal{B}^{-1}(Y)} \prod_t p_\mathrm{trans}(Z[t]|X_\mathrm{enc}, \mathcal{B}(Z[1:t-1])).
\end{aligned}
\end{equation}
It should be noted that the definitions of $\mathcal{B}$ and $\mathcal{B}^{-1}$ are slightly different from those of CTC.
However, the summation in the loss function can efficiently be computed via the forward-backward algorithm as similar to the case of CTC.

\subsection{Attention decoder}

Attention decoders~\cite{Chorowski2014} can be expressed in two modules: $\textrm{Attend}(\cdot, \cdot)$ and $\textrm{Spell}(\cdot, \cdot)$.
Unlike CTC and transducer, the attention decoder does not explicitly align sequences of speech and text. Instead, it uses an attention mechanism $\textrm{Attend}(X_\mathrm{enc}, Y[1:t-1])$ to directly summarize encoded speech and prefix text having different lengths into a fixed-dimensional context vectors $A$:
\begin{equation}
\begin{aligned}
    A[t] &= \textrm{Attend}(X_\mathrm{enc}, Y[1:t-1]).
\end{aligned}
\end{equation}
The predictive distribution over target tokens are defined with another network $\textrm{Spell}$, as follows:
\begin{equation}
\begin{aligned}
    p_\textrm{att}(Y[t] | X_\textrm{enc}, Y[1:t-1]) &= \textrm{Spell}(Y[1:t-1], A[t]).
\end{aligned} \label{eq:att_prediction}
\end{equation}

Using the prediction defined in Eq. \eqref{eq:att_prediction}, the loss function to be minimized in training can be expressed as follows:
\begin{equation}
\begin{aligned}
    L_\textrm{att} &= -\log \prod_t p_\textrm{att}(Y[t] | X_\textrm{enc}, Y[1:t-1]).
\end{aligned}
\end{equation}
This loss function is a straight-forward extension of the cross-entropy loss function, and therefore, it requires no forward-backward algorithm in the previously introduced decoders.

\section{Training techniques}
\label{sec:optimization}

\subsection{SpecAugment}

SpecAugment~\cite{SpecAugment} is a data augmentation method for ASR. It transforms speech time-frequency features by time masking and frequency masking \footnote{Time warping module introduced in~\cite{SpecAugment} is unused in this paper.}.
The time and frequency masking processes samples random segments along with the time and frequency axis, respectively.
Masking is done by setting input features in the sampled segments to be zero.

This process has four hyperparameters, $T_n, F_n, T_w, F_w$.
$T_n$ and $F_n$ are the numbers of segments to be sampled for time and frequency masking, respectively.
$T_w$ and $F_w$ are the maximum widths of the time and frequency segments, respectively.

\subsection{Exponential moving average}

To improve the stability and the generalization ability of the estimated parameters,
we apply exponential moving average (EMA) for computing the final estimation results from the optimization trajectory.

After each training step $k$, we estimate EMA of the network parameters $\theta'_k$ from the current parameter $\theta_k$~\cite{zhang2020pushing}:
\begin{equation}
\begin{aligned}
    \theta'_k = \gamma \theta'_{k-1} + (1 - \gamma) \theta_k,
\end{aligned}
\end{equation}
where $\gamma \in (0, 1)$ is a decay rate.

\subsection{Variational noise}

Variational noise (VN)~\cite{vn} approximates Bayesian inference to improve generalization ability of neural networks (NN). 
In this method, a noised parameter vector $\theta'$ is introduced in training.
The noised parameter vector $\theta'$ is obtained from the original parameter vector $\theta$ by adding noise samples $n$ drawn from a normal distribution, as follows:
\begin{equation}
\begin{aligned}
    \theta' = \theta + n, n \sim \text{Normal}(0, \sigma^2),
\end{aligned}
\end{equation}
where $\sigma$ is a standard deviation of noise.

\section{Experiments}

In this section, we first conducted the comparative experiments on different architectures and loss functions, and then conducted the ablation study using the best performing configuration.
The computational efficiency of the compared models was also compared and discussed.

\subsection{Corpus of Spontaneous Japanese}

CSJ is an ASR corpus consisting of 581 hours of speech and transcriptions.
The transcriptions were tokenized into character-based tokens consisting of all 3259 characters in the corpus augmented with 3 special tokens: Start-of-sentence (SOS), end-of-sentence (EOS), and unknown (UNK) special tokens.
The inputs for neural networks were computed by extracting features from the waveform of the utterances.

As speech features, 80-dimensional log mel-filterbank outputs were computed from 40ms window for each 10ms.
Those log mel-filterbank features were further normalized to have zero mean and unit variance over the training partition of the dataset.

It should be noted that CSJ does not define a separated development set. Therefore, in the experiments, a part of training set is used as a development set. We followed a rule employed in ESPnet for splitting the training set into training and development partitions so that the numbers reported in this paper can be directly compared to the results of ESPnet.

\subsection{Hyperparameter settings}
\label{sec:settings}

We set the sizes of neural networks to be similar to the best performing configuration found on LibriSpeech~\cite{librispeech} tasks because the scale of CSJ task is close to that of Librispeech task. %
For BLSTM models, we employed \textit{LAS-6-1280} in  \cite{SpecAugment} with 6 BLSTM encoder layers. And the attention and transducer decoder have one 1280-dim LSTM layer and 128-dim embedding. The attention decoder for the BLSTM encoder also consists of a 128-dim additive attention layer~\cite{attention}.

For our Conformer models, we selected \textit{Conformer L} from~\cite{Gulati2020}, which has 17 Conformer blocks of 512-dim 8-head dot attention, 512-dim 32-frame kernel 1d convolution, and 2048-dim feedforward module. The attention and transducer decoder consist of 640-dim LSTM layer and 128-dim embedding. The attention decoder additionally used a scaled 512-dim 8-head dot attention layer~\cite{transformer} for the Conformer encoder. 

Note that, every encoder network has two 2D convolutional input layers with kernel size 3 and stride 2 to subsample the speech features before Conformer blocks or BLSTM layers. %

For the configuration of training techniques in Section \ref{sec:optimization}, we set EMA decay rate $\gamma$ to 0.9999, and VN scale $\sigma$ to 0.075~\footnote{We applied VN only in embedding and LSTM layers}.
Hyperparameters of SpecAugment were  $(T_n, T_w, F_n, F_w) = (1, 50, 1, 15)$ for BLSTM encoders, and $(T_n, T_w, F_n, F_w) = (10, 0.05 \times T, 2, 27)$ for Conformer, respectively, where $T$ is the number of speech feature frames.
We configured Adam optimizer~\cite{kingma2014adam} with learning rate scheduling in~\cite{SpecAugment} for BLSTM models, and~\cite{Gulati2020} for Conformer models, respectively.
We trained all models up to 150k steps (i.e. 200 epochs) until convergence.
 
We implemented all our models using Lingvo~\cite{shen2019lingvo}.
During training, we used 32 chips of Google tensor processing unit v3~\cite{kumar2019scale} in parallel. Batch size per chip was 16/ 64 for Conformer/ BLSTM models, respectively.

In a decoding stage, the attention and transducer decoders kept 8 hypotheses in beam search, while CTC performed greedy search.
During search, we omitted EOS if its logit is less than 5.0 to prevent shorter hypotheses~\cite{Chorowski2017}, and used no language models.

\subsection{Comparison on character error rates}

\begin{figure}
\begin{center}
  \includegraphics[width=\linewidth]{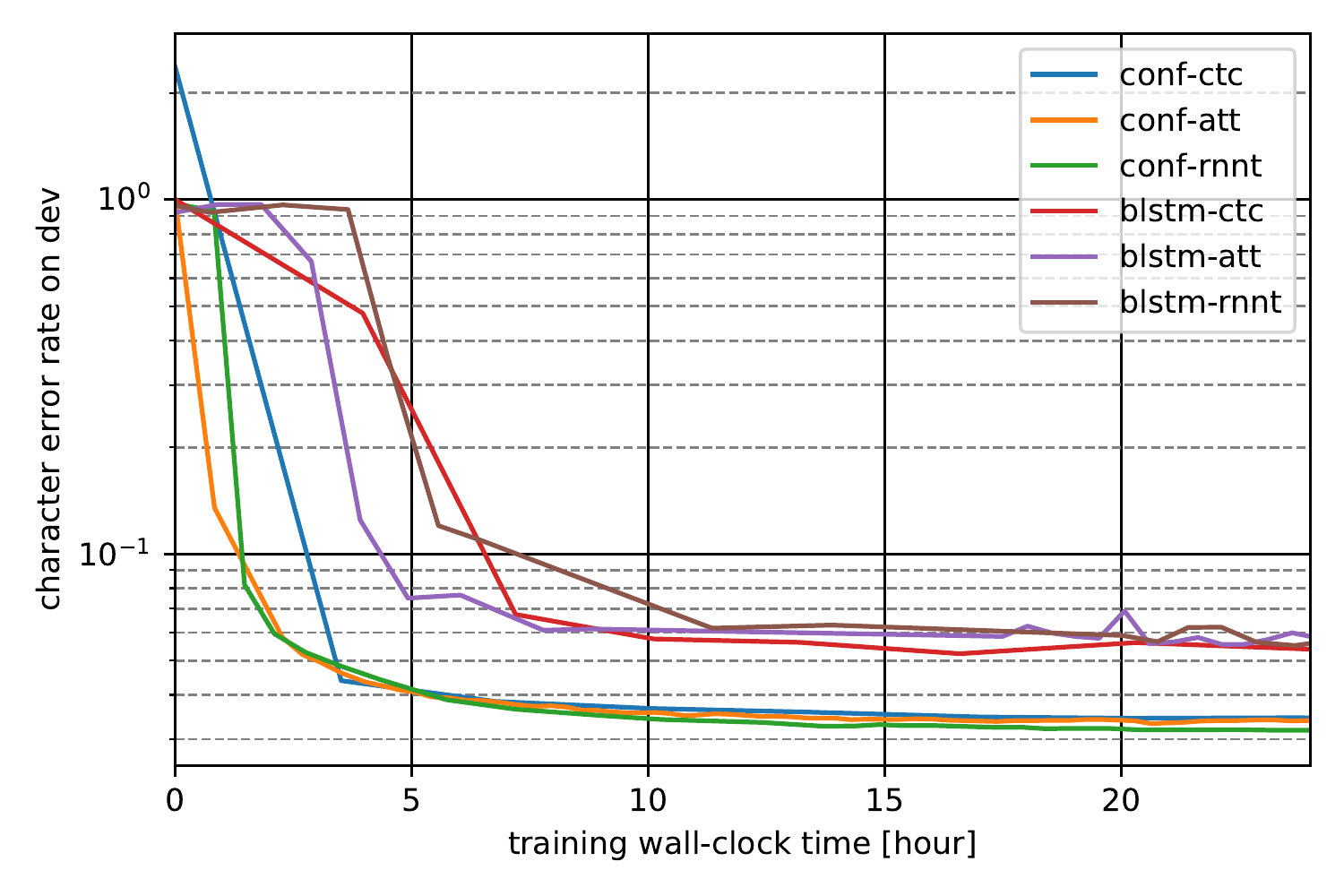}
\end{center}
\vspace{-0.8cm}
  \caption{Learning curves of character error rates on CSJ dev for each network architecture.}
  \label{fig:network-cer-curve}
\vspace{-0.5cm}
\end{figure}

We measured Japanese ASR performance in our systems using CER.
Table~\ref{tab:results} shows CERs on CSJ dev / eval1 / eval2 / eval3, respectively, and training throughput in utterances per second (utt/sec). 
Figure~\ref{fig:network-cer-curve} shows convergence of CERs over the development set as functions of training wall-clock time.

In Table~\ref{tab:results}, our Conformer transducer achieved the lowest CERs of 4.1\% / 3.2\% / 3.5\% on CSJ eval1 / eval2 / eval3, respectively.~\footnote{We also tested a Conformer wav2vec 2.0 encoder~\cite{zhang2020pushing} that has 0.6B parameters pretrained with Libri-Light 60K hour unsupervised English speech~\cite{librilight}. It did not reduce CERs from our Conformer transducer.}  Comparing encoder implementations, the Conformer encoder always achieved lower CERs than the BLSTM encoder. In contrast, differences in CERs between decoders were less significant than those between encoders.

To the best of our knowledge, the CERs of our Conformer transducer are lower than the existing reports on CSJ. For example, ESPnet2 Conformer model achieved CER 4.5\% / 3.3\% / 3.6\% in~\cite{guo2020recent}, and Transformer model~\cite{Moriya2020} reported CER 4.6\% / 3.4\% / 3.7\% on CSJ eval1 / eval2 / eval3, respectively.

On training convergence in Figure~\ref{fig:network-cer-curve}, Conformer encoder based systems, labeled ``conf-*'', clearly converged faster and showed lower CERs regardless of the choice of decoders.
Therefore, we could recommend the Conformer as a drop-in replacement of the BLSTM encoder.
Comparing the decoders, the attention decoders labeled ``*-att'' result in lower CERs than others at early steps but their differences were smaller than those of encoders.

\begin{table}[tb]
  \caption{Character error rates on CSJ dev / eval1 / eval2 / eval3 sets, and training throughput for each network architecture. %
  }
   \vspace{-0.3cm}
  \label{tab:results}
  \centering
  \footnotesize
  \begin{tabular}{ llrrr }
    \toprule
    \textbf{Encoder} & \textbf{Decoder} & \textbf{\#Param} & \textbf{Utt/sec} & \textbf{CER} [\%] \\
    \midrule
    BLSTM  & CTC & 258M & 430.0 & 3.9 / 5.2 / 3.7 / 4.0 \\
    BLSTM & attention & 309M & 365.5 & 3.8 / 5.3 / 3.7 / 3.7 \\
    BLSTM & transducer & 274M & 297.6 & 3.8 / 5.1 / 3.7 / 4.0 \\
    Conformer & CTC & 117M & \textbf{628.4} & 3.5 / 4.6 / 3.3 / 3.7 \\
    Conformer & attention & 124M & 534.8 & 3.3 / 4.6 / 3.4 / 3.6 \\
    Conformer & transducer & 120M & 376.1 & \textbf{3.1} / \textbf{4.1} / \textbf{3.2} / \textbf{3.5} \\
    \midrule
    \multicolumn{2}{l}{Moriya et al.~\cite{Moriya2020}} & - & - & - / 4.6 / 3.4 / 3.7 \\
    \multicolumn{2}{l}{Guo et al.~\cite{guo2020recent,kan_bayashi_2020_4065140}} & 91M & - & - / 4.5 / 3.3 / 3.6 \\
\bottomrule
  \end{tabular}
  \vspace{-0.5cm}
\end{table}

\subsection{Ablation study on training techniques}

\begin{figure}
\begin{center}
  \includegraphics[width=\linewidth]{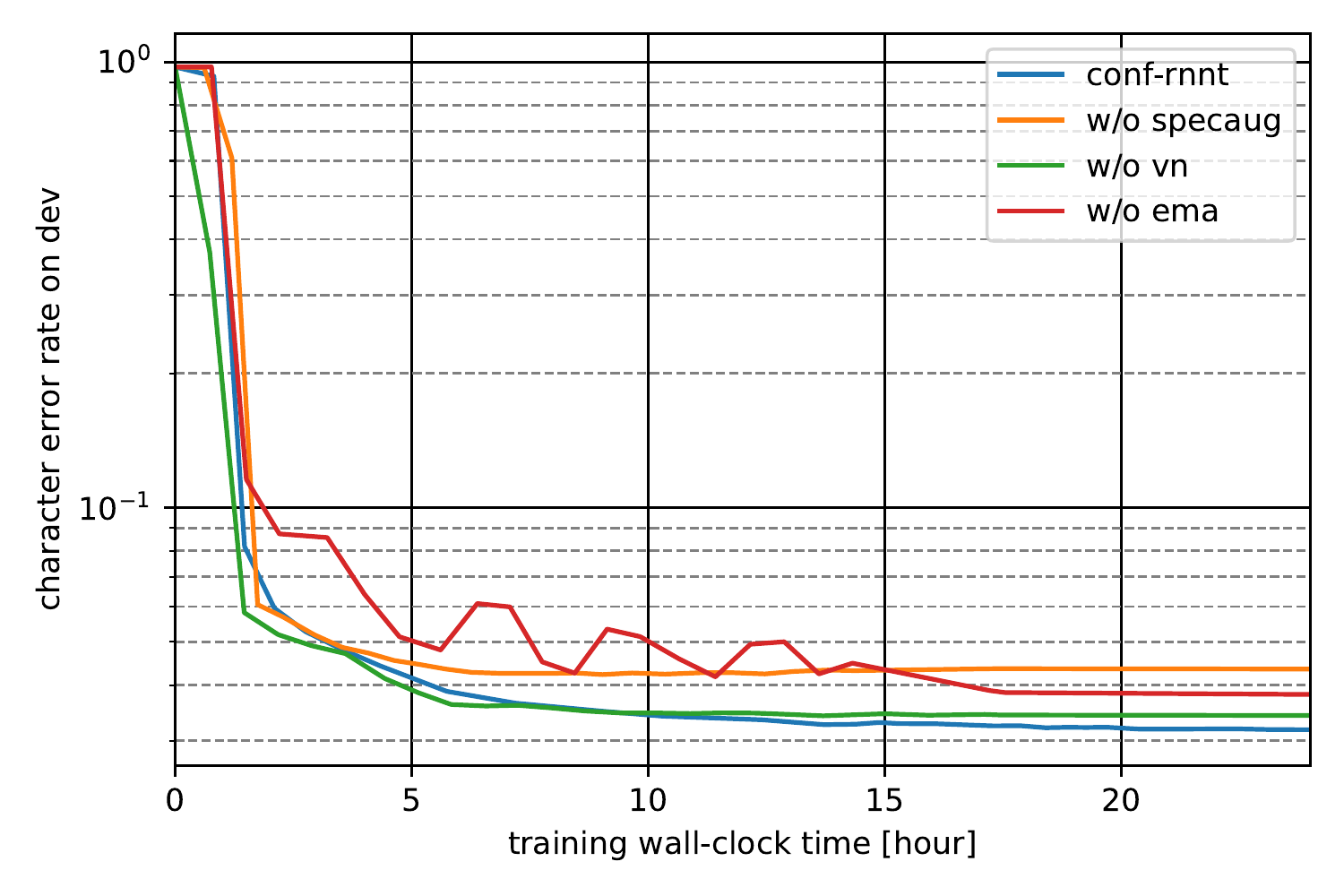}
\end{center}
\vspace{-0.8cm}
  \caption{Learning curves of character error rates on CSJ dev for each system in Table~\ref{tab:ablation}.}
  \label{fig:opt-cer-curve}
\end{figure}

To find which training techniques were effective in our best system, we removed components in Section~\ref{sec:optimization} one by one.
Table~\ref{tab:ablation} summarizes training throughput, and CER for each system that disabled one method. Figure~\ref{fig:opt-cer-curve} shows relationships between training wall-clock time and CER of systems in Table~\ref{tab:ablation}.

Comparing CERs in Table~\ref{tab:ablation}, SpecAugment was the most important training method because the system without it showed the worst dev CER 4.2\%. In addition, VN and EMA were essential to achieve the best performance as the system without VN/EMA degraded the dev CER to 3.4\%/3.7\%, respectively. Especially, in early steps, CERs of the system without EMA were unstable in Figure~\ref{fig:opt-cer-curve}.
In other words, EMA accelerated training convergence more than other methods. Therefore, these training methods were as important as the choice of encoder and decoder networks.

\begin{table}[tb]
\vspace{-0.3cm}
  \caption{Ablation study on our best Conformer transducer baseline. For example, ``w/o Variational noise'' denotes the system disabled VN but enabled SpecAugment (SA) and EMA.
  }
     \vspace{-0.3cm}
  \label{tab:ablation}
  \footnotesize
  \centering
  \begin{tabular}{ lrr }
    \toprule
     & \textbf{Utt/sec} & \textbf{CER} [\%] \\
    \midrule
    Baseline (w/ SA, VN, and EMA) & 376.1 & 3.1 / 4.1 / 3.2 / 3.5 \\
    w/o SpecAugment & 375.9 & 4.2 / 5.8 / 4.1 / 4.3 \\
    w/o Variational noise & 376.3 & 3.4 / 4.6 / 3.5 / 3.8 \\
    w/o Exponential moving average & 377.6 & 3.7 / 5.1 / 3.7 / 4.0 \\
    \bottomrule
  \end{tabular}
\end{table}

\subsection{Comparison on computational complexity}

Comparing encoder-decoder architectures in Table~\ref{tab:results}, Conformer CTC achieved the fastest training throughput of 628.4 utt/sec on the 32 TPUs. 
In other words, it took only 10.7 minutes to process all the training utterances in CSJ.
The efficiency came from the fact that Conformer CTC has no expensive recurrent operations as in LSTMs.
On the other hand, the combination of BLSTM encoder and transducer decoder was the slowest throughput of 297.6 utt/sec.
In Table~\ref{tab:ablation}, we also confirmed that CER and training speed degradation due to the training techniques introduced in this paper was negligible.

In order to compare inference-time computational efficiency, we measured the real-time factors (RTFs) of the decoding time with the compared models.
In RTF evaluation, accelerators such as TPUs are not used, and the decoding algorithm is executed using Intel(R) Xeon(R) Platinum 8273CL 2.20GHz CPU.
The decoding hyperparameters were set to be consistent with the training setting described in Section~\ref{sec:settings}, but we set batch size to be 1 in decoding.

Table~\ref{tab:rtf} lists the average and standard deviation (stddev)  of RTFs measured over the CSJ eval1 set.
From the table, the combination of Conformer encoder and transducer decoder shown to be more computationally efficient than other encoders and decoders.
This setting is identical with the setting used for obtaining the best character error rate in the previous section.

Since the beam search algorithm performed with neural E2E model is sufficiently computationally efficient, significant difference in inference speed is hardly obtained just by modifying the neural architecture.
However, in combination with the results of training speed shown in  Table~\ref{tab:results}, we could conclude that the combination of the Conformer and transducer is the most computationally efficient.

It should be noted that those results were better than the existing study on CSJ eval1 RTF of 0.09~\cite{Seki2019}.
\footnote{RTFs are not directly comparable because Lingvo implemented beam search in C++ while \cite{Seki2019} implemented them purely in Python.}

\begin{table}[tb]
\vspace{-0.3cm}
    \centering
    \caption{Real-time factor (RTF) on the CSJ eval1 set.}
     \vspace{-0.3cm}
    \begin{tabular}{llrr}
    \toprule
    \textbf{Encoder} & \textbf{Decoder} & \textbf{RTF} & \textbf{Stddev} \\
    \midrule
    BLSTM & CTC & 0.050 & 0.070 \\
    BLSTM & attention & 0.050 & 0.069 \\
    BLSTM & transducer & 0.048 & 0.068 \\
    Conformer & CTC & 0.047 & 0.066 \\
    Conformer & attention & 0.049 & 0.069 \\
    Conformer & transducer & \textbf{0.046} & \textbf{0.065} \\
    \bottomrule
    \end{tabular}
    \label{tab:rtf}
    \vspace{-0.5cm}
\end{table}

\section{Conclusions}

In this paper, 
we compared the latest network architectures with the advanced training methods for Japanese speech recognition.
As a result, it is found that combining Conformers with transducer decoder augmented by several training techniques achieved the best results in CSJ experiments.
The model introduced in this paper achieved the state-of-the art performance that are CERs of 4.1\% / 3.2\% / 3.5\% on eval1 / eval2 / eval3, respectively.
The model also shown to be more computationally efficient in decoding compared to the conventional models.

\bibliographystyle{IEEEtran}

\bibliography{main}

\end{document}